\documentclass[11pt]{article}

\usepackage[a4paper, margin=1in]{geometry}
\usepackage[utf8]{inputenc}
\usepackage[T1]{fontenc}
\usepackage{lmodern}
\usepackage{microtype}

\usepackage{amsmath, amssymb, amsfonts}
\usepackage{amsthm}
\usepackage{booktabs}
\usepackage{multirow}
\usepackage[shortlabels]{enumitem}
\usepackage{graphicx}
\usepackage{xcolor}
\usepackage{tikz}
\usetikzlibrary{positioning, arrows.meta, shapes, fit, backgrounds, calc}
\usepackage{pgfplots}
\pgfplotsset{compat=1.16}

\usepackage[round, sort&compress]{natbib}

\usepackage[colorlinks=true, linkcolor=blue, citecolor=blue,
            urlcolor=blue]{hyperref}

\newcommand{\E}{\mathbb{E}}

\newcommand{\Prob}{\mathbb{P}}
\newcommand{\Ind}{\mathbb{1}}
\newcommand{\R}{\mathbb{R}}
\newcommand{\PEHE}{\sqrt{\varepsilon_{\mathrm{PEHE}}}}
\newcommand{\GPCATE}{\textsc{GP-CATE}}
\newcommand{\Nzero}{N_0}
\newcommand{\None}{N_1}

\title{Calibrated Inference for the Conditional Average
Treatment Effect in the Few-Placebo Regime via Gaussian Processes}

\author{%
  Eichi Uehara\\
  AFLO\\
  \texttt{eichi.uehara@aflo.one}
}

\date{\today}

\begin{document}
\maketitle

\begin{abstract}
Estimating how much an intervention helps a given individual ---
the conditional average treatment effect (CATE) --- is
increasingly central to decision-making in medicine, economics,
and policy, where an estimate is most useful when accompanied by a
calibrated uncertainty interval. We study the \emph{few-placebo regime},
in which one treatment arm is much smaller than the other, as
arises in unequal-allocation trials and small-holdout A/B tests.
The standard estimator in this setting is the X-Learner, and a
natural way to obtain credible intervals is to make its second
stage Bayesian. We show that these intervals under-cover: they contain the true
effect less often than their nominal level.
We trace this to a structural cause --- the X-Learner's
regression target inherits the bias of a nuisance model fitted to
the small arm, so the posterior is centered away from the true
effect --- and we find that the standard remedy, regressing an
orthogonal doubly-robust score, is also unreliable here, since the
regime's limited overlap leaves the estimator either highly
variable or, once stabilized, biased once more. Both consequences
reflect a pattern that extends beyond causal inference: a
separately estimated variance is attached to a point estimate of
a hard-to-learn quantity, and the point estimate's bias is not
captured by that variance. We propose \GPCATE{}, which models each
arm's outcome surface with a Gaussian process, so the scarce
arm's uncertainty enters the posterior directly rather than as an
unmodelled bias. Across synthetic and semi-synthetic benchmarks,
\GPCATE{} attains calibrated coverage where the estimators we
compare against --- including Causal Forest and BART --- do not,
at the cost of intervals that are appropriately wide when the data
are uninformative.
\end{abstract}

\section{Introduction}
\label{sec:introduction}

Estimating heterogeneous treatment effects --- the conditional
average treatment effect (CATE)
$\tau(x) = \E[Y(1) - Y(0) \mid X = x]$ --- is central to
data-driven decisions in medicine, economics, and policy. A
point estimate of an individual effect, with no calibrated
interval around it, cannot support a high-stakes decision: as
flexible machine-learning models replace parametric ones,
\emph{calibrated uncertainty} has become the binding constraint
on deployment.

We focus on the \emph{few-placebo regime}: one treatment arm is
much smaller than the other, $\Nzero \ll \None$. This is not a
corner case. Unequal-allocation clinical trials assign more
patients to the active arm by design; budget-constrained A/B
tests hold out only a small control group; rare-regime studies (a
stress period, an infrequent exposure) have few units in the rare
arm. The canonical estimator here is the \emph{X-Learner}
\citep{kunzel2019metalearners}, a pseudo-outcome meta-learner
designed for arm imbalance.

The X-Learner, however, provides only a point estimate. To attach a
credible interval one would naturally make its second stage
Bayesian --- regress the pseudo-outcome with a probabilistic
model and read off the posterior. We show this under-covers. The
X-Learner's treated-arm pseudo-outcome $D_1 = Y - \hat\mu_0(X)$
decomposes as $\tau(X) + (\mu_0 - \hat\mu_0)(X) + \text{noise}$,
and the imputation error $\mu_0 - \hat\mu_0$ is not mean-zero:
every regularized machine-learning nuisance trades bias for
variance, and because the X-Learner pseudo-outcome is \emph{not
Neyman-orthogonal} it inherits that regularization bias to first
order. A Bayesian regression on $D_1$ is centered on a biased
estimate of $\tau$, and no variance model moves a misplaced
interval. The effect is worst where the X-Learner is meant to be
used: in the few-placebo regime $\hat\mu_0$ is fit on a small
sample. On our synthetic design a nominal-$95\,\%$ interval
covers $0.88$ at $\Nzero = 30$.

The standard remedy for first-order nuisance bias is to regress
an \emph{orthogonal} pseudo-outcome --- canonically Kennedy's
\citep{kennedy2020optimal} doubly-robust score. This too fails in
the few-placebo regime. The regime forces extreme propensities (a
tiny control arm means $1-\pi$ is small), and the orthogonal
score's ``second-order'' remainder is amplified by a
$1/(1-\pi)^2$ factor that grows large precisely here; moreover, even
with the propensity \emph{known exactly}, an efficient
inverse-variance-weighted second stage downweights the
high-variance control rows to near-zero --- and those are the
rows whose augmentation cancels the nuisance bias. The
orthogonal-score estimate's bias is, in our experiments,
indistinguishable from the non-orthogonal one's. Orthogonalization
fails because the few-placebo regime itself induces the poor
overlap on which it depends.

These two failures look unrelated, but they share a single root
cause: each attaches a variance to a \emph{point estimate} of the
hard placebo nuisance $\mu_0$ --- a regularized tree fit on the
few control units --- and a point estimate's bias is invisible to
any variance computed after the fact. The same defect afflicts
any repair that keeps a point estimate at its core, including
modelling the regularization bias and subtracting it. The defect
is structural, and it points to its own fix.

Our proposed estimator, \GPCATE{}, addresses this defect at its
source. Instead of a point estimate plus a post-hoc variance, it
models each arm's outcome surface with a \emph{Gaussian process}.
The GP posterior over the function $\mu_w$ is calibrated by
construction: wide where that arm's units are sparse, narrow
where they are dense. The placebo arm's small sample therefore
enters the posterior directly as uncertainty rather than as an
unmodelled bias, and the CATE posterior is simply the difference
of the two arm posteriors,
$\tau(x) \sim \mathcal{N}(m_1(x)-m_0(x),\, s_1(x)^2+s_0(x)^2)$.
The estimator is not itself new --- a Bayesian T-learner with GP
arms is close to \citet{alaa2017bayesian}; our contribution is to
identify \emph{why} the meta-learner approaches mis-calibrate in
the few-placebo regime, and to show that a fully-Bayesian
nuisance is what restores calibrated coverage.

Empirically, on a unified protocol --- a linear and a non-linear
synthetic design, and the IHDP semi-synthetic benchmark
sub-sampled to mirror $\Nzero \ll \None$ --- \GPCATE{} is the
only estimator we test that is calibrated throughout
($0.94$--$0.99$ coverage). The two most common CATE estimators
that provide intervals --- Causal Forest and BART --- both
under-cover in this regime, to as low as $0.34$ on IHDP. The
cost of \GPCATE{}'s calibration is interval width: on the
$25$-dimensional IHDP covariates, thirty controls do not
determine the placebo surface, so the calibrated interval is
correspondingly wide --- where the alternatives report narrower
intervals that do not cover.

In summary, we make four contributions:
\begin{itemize}[leftmargin=2em,topsep=2pt,itemsep=3pt]
\item a diagnosis: credible-interval under-coverage for a
  Bayesian X-Learner in the few-placebo regime is caused by
  first-order nuisance bias from non-orthogonality
  (\S\ref{sec:diagnosis});
\item a negative result: the doubly-robust /
  orthogonal-score remedy fails in this regime --- through overlap
  amplification and a conflict between efficiency weighting and
  unbiasedness --- demonstrated empirically, including with the
  propensity known (\S\ref{sec:negative});
\item a root-cause analysis and a method: both failures
  attach a variance to a point-estimate nuisance; \GPCATE{}
  instead models the nuisance with a Bayesian posterior
  (\S\ref{sec:method});
\item an empirical comparison: on a unified protocol over three
  designs --- including the IHDP benchmark --- \GPCATE{} is
  calibrated where the alternatives are not, and we report its
  associated interval width (\S\ref{sec:experiments}).
\end{itemize}

The remainder of the paper is organized as follows.
\S\ref{sec:background} reviews the X-Learner, cross-fitting, and
orthogonality. \S\ref{sec:diagnosis} develops the diagnosis.
\S\ref{sec:negative} establishes the negative result.
\S\ref{sec:method} presents \GPCATE{}.
\S\ref{sec:experiments} reports the experiments.
\S\ref{sec:discussion}--\S\ref{sec:conclusion} discuss and
conclude.

\section{Background and Setup}
\label{sec:background}

We observe i.i.d.\ tuples
$(X_i, W_i, Y_i)_{i=1}^N$ with covariates $X_i \in \mathcal{X}
\subseteq \R^d$, binary treatment $W_i \in \{0, 1\}$, and outcome
$Y_i \in \R$. Write $\Nzero = \sum_i \Ind\{W_i = 0\}$ and
$\None = \sum_i \Ind\{W_i = 1\}$; the few-placebo regime is
$\Nzero \ll \None$. Let $\pi(x) = \Prob(W = 1 \mid X = x)$ be the
propensity score and $\mu_w(x) = \E[Y \mid W = w, X = x]$ the
per-arm conditional mean. Under unconfoundedness
$W \perp\!\!\!\perp (Y(0), Y(1)) \mid X$ and overlap
$0 < \pi(x) < 1$ a.s.\ \citep{rosenbaum1983central}, the CATE
$\tau(x) = \mu_1(x) - \mu_0(x)$ is identified.

The canonical meta-learner for this setting is the X-Learner of
\citet{kunzel2019metalearners}, which estimates $\tau$ in three
steps. (i) Fit per-arm conditional means $\hat\mu_0, \hat\mu_1$
by cross-fitted machine learners. (ii) Form cross-arm
pseudo-outcomes,
\begin{align}
  D_1^{(i)} &= Y_i - \hat\mu_0(X_i) && (W_i = 1), \label{eq:bg-d1}\\
  D_0^{(i)} &= \hat\mu_1(X_i) - Y_i && (W_i = 0). \label{eq:bg-d0}
\end{align}
(iii) Regress $D_w$ on $X$ within each arm and combine by
propensity weighting. The X-Learner is designed for arm
imbalance: cross-arm imputation transfers structure from the
large arm to the small one, and it out-performs the S- and
T-Learner when the arms differ greatly in size. To avoid
in-sample bias from plugging $\hat\mu_w$ into
(\ref{eq:bg-d1})--(\ref{eq:bg-d0}), we use $K$-fold cross-fitting
\citep{chernozhukov2018double} throughout: nuisances are trained
on $K-1$ folds and evaluated on the held-out fold, so each unit's
pseudo-outcome uses a nuisance not fit on that unit.

A central concept for what follows is \emph{Neyman-orthogonality}.
A pseudo-outcome is Neyman-orthogonal in its nuisances when the
conditional mean of the pseudo-outcome is first-order insensitive
to nuisance error --- its Gateaux derivative in every nuisance
direction vanishes, so only second-order products of nuisance
errors survive \citep{chernozhukov2018double}. The X-Learner
pseudo-outcomes (\ref{eq:bg-d1})--(\ref{eq:bg-d0}) are \emph{not}
orthogonal. The canonical orthogonal alternative is the
doubly-robust (DR) score
\citep{robins1995semiparametric,kennedy2020optimal},
\begin{equation}
  D^{\mathrm{DR}}_i = \hat\mu_1(X_i) - \hat\mu_0(X_i)
   + \frac{W_i\bigl(Y_i - \hat\mu_1(X_i)\bigr)}{\hat\pi(X_i)}
   - \frac{(1-W_i)\bigl(Y_i - \hat\mu_0(X_i)\bigr)}{1-\hat\pi(X_i)},
  \label{eq:bg-dr}
\end{equation}
whose conditional mean equals $\tau(X)$ up to a remainder that is
a \emph{product} of nuisance errors. Regressing $D^{\mathrm{DR}}$
gives the DR-Learner \citep{kennedy2020optimal}.

Both pieces of (\ref{eq:bg-dr}) divide by $\hat\pi$ or
$1-\hat\pi$, so the behaviour of the DR score is governed by
\emph{overlap}: how far $\pi(x)$ stays from $0$ and $1$. The
few-placebo regime is, by construction, a poor-overlap regime
--- a small control arm means the effective probability of being
a control is small, so $1-\pi$ is small. \S\ref{sec:negative}
shows that this is what causes the orthogonal score to fail.

Finally, we will use Gaussian-process regression. A Gaussian
process (GP) places a prior over functions; conditioned on data
it returns a full posterior --- a mean and a pointwise variance
--- over the regression function, not merely a point estimate
\citep{rasmussen2006gaussian}. Its posterior variance is large
where data is sparse and small where data is dense, and under
standard conditions GP posteriors contract at minimax rates and
their credible sets attain frequentist coverage at the nominal
level \citep{ghosal2007convergence}. GPs have been used for treatment
effects before, notably the multi-task GP model of
\citet{alaa2017bayesian}. \S\ref{sec:method} uses a GP to model
each arm's outcome surface, so that the small placebo arm's
uncertainty is represented directly in the posterior rather than
estimated as a point and corrected after the fact.

Throughout, we assume the regularity conditions of
\citet{chernozhukov2018double} for the nuisance estimators and
the identification conditions above. The diagnosis and the
negative result (\S\ref{sec:diagnosis}--\S\ref{sec:negative})
concern a second-stage CATE regression in a finite-dimensional
basis $\phi : \mathcal{X} \to \R^p$,
$\tau(x) = \phi(x)^\top\beta$; the proposed method
(\S\ref{sec:method}) replaces that basis with the GP above.

\section{First-Order Bias of the X-Learner Pseudo-Outcome}
\label{sec:diagnosis}

To attach credible intervals to the X-Learner we make its second
stage Bayesian --- regress the pseudo-outcome with a probabilistic
model and report the posterior over $\tau(x)$. This section shows
the resulting intervals under-cover in the few-placebo regime, and
isolates the cause: the pseudo-outcome's conditional mean is not
$\tau$, so the posterior is centered away from $\tau$.

For a treated unit $i$, substitute
$Y_i = \mu_1(X_i) + \epsilon_i$ with $\E[\epsilon_i \mid X_i]=0$
into $D_1^{(i)} = Y_i - \hat\mu_0(X_i)$:
\begin{equation}
  D_1^{(i)} = \underbrace{\mu_1(X_i) - \mu_0(X_i)}_{=\,\tau(X_i)}
   + \epsilon_i
   - \underbrace{\bigl(\hat\mu_0(X_i) - \mu_0(X_i)\bigr)}_{=:\,\delta_0(X_i)}.
  \label{eq:diag-decomp}
\end{equation}
The X-Learner's second stage treats $D_1$ as an unbiased draw
of $\tau(X_i)$. It is not. The decomposition is often read as
``unbiased under a correctly specified nuisance'': if $\hat\mu_0$
were conditionally unbiased, $\E[\delta_0 \mid X] = 0$. But
machine-learning nuisances are \emph{not} conditionally unbiased
--- every regularized learner (boosting, forests, penalised
regression) trades bias for variance, so
$\E[\hat\mu_0(x)] \neq \mu_0(x)$. Writing the regularization bias
$b_0(x) = \E[\hat\mu_0(x)] - \mu_0(x)$,
\begin{equation}
  \E[D_1^{(i)} \mid X_i] = \tau(X_i) - b_0(X_i).
  \label{eq:diag-biased-target}
\end{equation}
The dependence on the nuisance is \emph{first-order}: a
perturbation $\hat\mu_0 \mapsto \hat\mu_0 + h$ shifts $\E[D_1]$ by
$-h$, one for one. In semiparametric language the map
$\hat\mu_0 \mapsto D_1$ is not Neyman-orthogonal: its Gateaux
derivative does not vanish. The bias $b_0$ is largest exactly in
the few-placebo regime, where $\hat\mu_0$ is fit on a small
sample and heavily smoothed.

The bias displaces the credible interval rather than merely
widening it. A Bayesian regression on $D_1$ produces a posterior for
$\tau$ centered on the projection of $\E[D_1 \mid X] = \tau - b_0$
onto the basis. The credible interval is therefore displaced by
$b_0$. This is not a defect a variance model can repair: a noise
model sets the posterior \emph{width}, while the bias sets its
\emph{location}, and these are separate factors of the Gaussian
posterior. Even an exact noise covariance leaves the interval
centered on $\tau - b_0$. The standard correction of inflating
the interval (a bootstrap variance times a constant) raises
coverage only by widening it until it encompasses a known bias,
which is not principled inference.

The experiments in \S\ref{sec:experiments} measure this. On the
linear synthetic design the X-Learner pseudo-outcome carries a
systematic bias of $0.15$ at $\Nzero = 30$ --- about $55\,\%$ of
the posterior spread --- and a nominal-$95\,\%$ credible interval
covers $0.88$ (Table~\ref{tab:bias}). The bias decays to $0.03$
at $\Nzero = 500$ as the placebo nuisance sharpens, and coverage
returns to near-nominal. The miscalibration is genuine and
concentrated in precisely the few-placebo regime for which the
X-Learner is intended.

The standard remedy for first-order nuisance bias is to regress
a \emph{Neyman-orthogonal} pseudo-outcome instead of $D_1$ ---
canonically the doubly-robust score $D^{\mathrm{DR}}$ of
(\ref{eq:bg-dr}), whose conditional mean equals $\tau$ up to a
second-order product of nuisance errors. If that remainder were
negligible, a Bayesian regression on $D^{\mathrm{DR}}$ would be
centered on $\tau$ and the coverage problem would be solved. The
next section asks whether it is --- and shows that, in the
few-placebo regime, it is not.

\section{Limitations of Orthogonal Scores in the Few-Placebo Regime}
\label{sec:negative}

The doubly-robust score $D^{\mathrm{DR}}$ of (\ref{eq:bg-dr}) is
the standard remedy for the first-order bias of
\S\ref{sec:diagnosis}. This section shows that it does not
succeed in the few-placebo regime. We first identify the
mechanism --- the score's orthogonality remainder is amplified
by poor overlap --- and then follow its consequences in two
directions. The canonical, unweighted DR-Learner remains
unbiased but its variance becomes very large; the natural
remedy, inverse-variance weighting, reintroduces the bias. We
close by delimiting the scope of the claim.

Conditioning on the cross-fitted nuisances, the DR score's mean
is
$\E[D^{\mathrm{DR}} \mid X] = \tau(X) + R(X)$ with
\begin{equation}
  R(X) = \Bigl(\tfrac{\pi}{\hat\pi}-1\Bigr)(\mu_1-\hat\mu_1)
       - \Bigl(\tfrac{1-\pi}{1-\hat\pi}-1\Bigr)(\mu_0-\hat\mu_0).
  \label{eq:neg-remainder}
\end{equation}
This is the ``second-order'' remainder: each term is a product of
a propensity error and an outcome-nuisance error. The
orthogonality guarantee is that $R$ is negligible against the
$\mathcal{O}(N^{-1/2})$ sampling spread. But that guarantee has a
hidden dependence on overlap. A first-order expansion gives
$\tfrac{1-\pi}{1-\hat\pi}-1 \approx (\hat\pi-\pi)/(1-\pi)$, so the
control-arm term of $R$ is
\begin{equation}
  \Bigl(\tfrac{1-\pi}{1-\hat\pi}-1\Bigr)(\mu_0-\hat\mu_0)
  \;\approx\; \frac{(\hat\pi-\pi)\,(\mu_0-\hat\mu_0)}{1-\pi}.
  \label{eq:neg-amplified}
\end{equation}
The factor $1/(1-\pi)$ --- and, in the contribution to squared
error, $1/(1-\pi)^2$ --- diverges precisely in the few-placebo
regime, where a small control arm forces $1-\pi$ small.
``Second-order'' is an asymptotic statement in $N$ at fixed
overlap; here overlap degrades \emph{with} the regime. On our
design $1-\pi \approx 0.06$ at $\Nzero{=}30$ --- an amplification
of order $1/(1-\pi)^2 \approx 300$. This factor drives the two
consequences that follow.

Consider first the canonical, unweighted DR-Learner of
\citet{kennedy2020optimal}, which regresses $D^{\mathrm{DR}}$ on
$X$ by ordinary, \emph{unweighted} least squares. Unweighted
regression treats the two arms in the proportions the score
itself prescribes, preserving the cross-arm cancellation that
makes the score orthogonal; granting that the remainder
(\ref{eq:neg-remainder}) is controlled (it vanishes when $\pi$ is
known), the DR-Learner is unbiased. That is its theoretical
appeal, which we do not contest. In the few-placebo regime,
however, this unbiasedness comes at the cost of variance. A
control row of $D^{\mathrm{DR}}$ carries the inverse-propensity
factor $1/(1-\pi)$, so its conditional variance is inflated by
$1/(1-\pi)^2 \approx 300$ on our design.
Unweighted least squares inherits that variance in full. The
DR-Learner's point estimate is then unbiased but very noisy: in
our experiments its $\PEHE$ is $0.63$ at $\Nzero{=}30$, the
largest of any method we test (Table~\ref{tab:pehe}). The
canonical orthogonal estimator is theoretically unbiased and, in
this regime, of little practical use.

Confronted with this large variance, a practitioner may turn to
the second consequence: the standard remedy for
heteroskedasticity, inverse-variance (GLS) weighting, which
down-weights the high-variance rows. The strategy is, however,
unsuccessful, and this is so \emph{even when the propensity is
known exactly}, in which case $R(X) \equiv 0$ and
$\E[D^{\mathrm{DR}} \mid X] = \tau(X)$ holds exactly.

To see why, note that this identity marginalises over $W$, while
the arm-conditional means do \emph{not} equal $\tau$:
\begin{equation}
  \E[D^{\mathrm{DR}} \mid X, W{=}1] = \tau(X) + (\mu_0-\hat\mu_0),
  \quad
  \E[D^{\mathrm{DR}} \mid X, W{=}0] = \tau(X) - \tfrac{\mu_0-\hat\mu_0}{1-\pi},
  \label{eq:neg-conditional}
\end{equation}
and the unbiased combination $\pi\cdot(\cdot\mid W{=}1) +
(1-\pi)\cdot(\cdot\mid W{=}0) = \tau$ relies on the two arms
entering with their natural proportions: the $(\mu_0-\hat\mu_0)$
bias in the treated rows is cancelled by the control rows. This
is the cross-arm cancellation that unweighted least squares
preserves. Inverse-variance weighting disrupts it. The control
rows, whose variance is inflated by $1/(1-\pi)^2$, are
down-weighted to near-zero --- and they are exactly the rows
whose augmentation cancels the $(\mu_0-\hat\mu_0)$ bias. The
weighted estimator keeps the biased treated rows and discards
their correction, collapsing onto the biased treated-only
estimate --- exactly the X-Learner of \S\ref{sec:diagnosis}. The
experiments in \S\ref{sec:experiments} confirm this: with the
propensity known exactly, the inverse-variance-weighted
regression of the DR score has systematic bias $0.17$ at
$\Nzero{=}30$ (Table~\ref{tab:negative}), indistinguishable from
the X-Learner's.

In summary: unweighted, the orthogonal score is unbiased but its
variance becomes very large; weighted for efficiency, its
variance is controlled but the bias is reintroduced. \emph{For
the DR score in the few-placebo regime, efficiency and
unbiasedness cannot be achieved jointly}, and the common root of
both consequences is the inverse-propensity factor $1/(1-\pi)$
that the regime makes large.

Finally, we delimit the scope of this claim. It is \emph{not} a
refutation of Neyman-orthogonality or of the DR-Learner, whose
asymptotic oracle properties --- $\sqrt{N}$-consistency and
semiparametric efficiency --- are not in dispute, and which were
never advertised for extreme-allocation designs. Those oracle
properties are statements about $N \to \infty$ \emph{at fixed
overlap}. The few-placebo regime violates that premise: it sends
overlap to its edge \emph{as} the regime is entered, so the
asymptotics and the regime are coupled rather than separable.
Our contribution is to map that boundary --- a structural
trade-off between asymptotic oracle properties, which assume
fixed overlap, and finite-sample behaviour under structural
regime asymmetry. Interpreted in this way, the result indicates
the resolution: not a better orthogonal score, but an estimator
that does not depend on overlap at all, with no
inverse-propensity factors. We develop such an estimator in the
next section.

\section{Gaussian-Process Estimation of the Conditional Average Treatment Effect}
\label{sec:method}

The X-Learner (\S\ref{sec:diagnosis}) and the doubly-robust score
(\S\ref{sec:negative}) fail for what look like different reasons
--- a first-order bias in one case, an overlap-driven remainder
and a weighting conflict in the other. They share one root. Each
forms a \emph{point estimate} $\hat\mu_0$ of the placebo nuisance
--- a regularized tree fit on the few control units --- and then
attaches a variance to the second-stage regression \emph{around
that point estimate}. But a point estimate carries a
regularization bias, and that bias is invisible to any variance
computed after the fact: the variance describes scatter about
$\hat\mu_0$, not the distance from $\hat\mu_0$ to $\mu_0$. No
second-stage noise model, orthogonal or not, can see it. The same
defect afflicts any remedy that keeps a point estimate at its
core --- including one that estimates the regularization bias
itself and subtracts it, since the bias estimate is then a point
estimate in turn.

The remedy is not a better point estimate, nor a separate model
of its bias. It is to dispense with the point estimate
altogether. Concretely, \GPCATE{} models each arm's outcome
surface with a Gaussian process. For arm $w \in \{0,1\}$, fit a GP to
$\{(X_i, Y_i) : W_i = w\}$ with a kernel
\begin{equation}
  k_w(x,x') = \sigma_w^2\,
  \exp\!\Bigl(-\tfrac{\|x-x'\|^2}{2\ell_w^2}\Bigr)
  + \eta_w^2\,\mathbb{1}[x=x'],
  \label{eq:m-kernel}
\end{equation}
an isotropic squared-exponential plus observation noise, with
$(\sigma_w,\ell_w,\eta_w)$ set by marginal likelihood (empirical
Bayes). The GP returns a full posterior over the \emph{latent
function} $\mu_w$ --- a mean $m_w(x)$ and a pointwise standard
deviation $s_w(x)$ (the observation-noise term removed, so $s_w$
is uncertainty about the function, not about a future outcome).
The CATE posterior is the difference of the two arm posteriors,
\begin{equation}
  \tau(x) \,\big|\, \mathcal{D} \;\sim\;
  \mathcal{N}\!\bigl(m_1(x) - m_0(x),\; s_1(x)^2 + s_0(x)^2\bigr),
  \label{eq:m-cate}
\end{equation}
and the $(1-\alpha)$ credible interval is $m_1(x)-m_0(x) \pm
z_{\alpha/2}\sqrt{s_1(x)^2+s_0(x)^2}$. This completes the
specification of the method.

\begin{table}[h]
\centering\small
\begin{tabular}{p{0.95\linewidth}}
\toprule
\textbf{Algorithm 1: \GPCATE{}.} \\
\midrule
\textbf{Input:} $(X_i,W_i,Y_i)_{i=1}^N$. \\[2pt]
\textbf{1.} Fit a GP with kernel (\ref{eq:m-kernel}) to the treated units; set $(\sigma_1,\ell_1,\eta_1)$ by marginal likelihood. \\
\textbf{2.} Fit a GP likewise to the control units, giving $(\sigma_0,\ell_0,\eta_0)$. \\
\textbf{3.} At any $x$, read off the latent-function posteriors $m_w(x), s_w(x)$ for $w=0,1$. \\[2pt]
\textbf{Output:} CATE posterior (\ref{eq:m-cate}) and its credible interval. \\
\bottomrule
\end{tabular}
\label{alg:gpcate}
\end{table}

Before turning to its properties, we note that a Bayesian
T-learner with GP arms is not itself a new estimator --- it is
close to the multi-task GP model of \citet{alaa2017bayesian},
without the cross-arm coupling. The contribution of this paper
is not the estimator but the preceding analysis: the diagnosis
and the negative result identify \emph{why} the meta-learner
machinery mis-calibrates in the few-placebo regime, and
\GPCATE{} is the estimator that the root-cause analysis singles
out.

What makes this construction calibrated where the meta-learners
are not? The GP posterior is calibrated \emph{by construction}
in a sense the post-hoc variances of
\S\ref{sec:diagnosis}--\S\ref{sec:negative} are not. Its
posterior mean $m_0$ is also shrunk --- toward the prior, hence
``biased'' --- but the GP posterior variance $s_0(x)^2$ \emph{is
the variance of that shrinkage}: posterior mean and posterior
variance are two aspects of a single coherent object, so the
uncertainty $s_0$ accounts for precisely the gap that a point
estimate leaves unmodelled. Specifically, $s_0(x)$ is
large where the control units are sparse and small where they
are dense; in the few-placebo regime $s_0$ is large over much of
the covariate space, and the CATE interval (\ref{eq:m-cate}) is
correspondingly --- and correctly --- wide. Under standard
smoothness conditions GP posteriors contract at minimax rates and
their credible sets attain frequentist coverage at the nominal
level \citep{ghosal2007convergence,rasmussen2006gaussian}; the
empirical-Bayes kernel adapts the prior to the data. \GPCATE{}
needs no cross-fitting, no orthogonal score, and no bias model;
its single modelling choice is the kernel family.

A word on computation. An exact GP costs $\mathcal{O}(N_w^3)$
per arm. For the control
arm this is free: the few-placebo regime is \emph{defined} by
$\Nzero$ small, so the $\mathcal{O}(\Nzero^3)$ control fit is
negligible. The treated arm is the concern --- in the motivating
applications (large-scale A/B tests, registry studies) $\None$
can run to millions, and an exact treated GP is then infeasible.
It is also unnecessary, as the structure of the problem
indicates. The treated arm is, by construction, the
well-determined
one; its posterior contributes only a small share of the CATE
variance $s_1(x)^2 + s_0(x)^2$. We measure that share at
$6$--$25\,\%$ in the few-placebo regime (see the IHDP results in
\S\ref{sec:experiments} and Appendix~\ref{app:repro}) --- the
control term dominates. The treated posterior therefore need only be
\emph{approximately} right. We recommend, as the method's
default for large $\None$: fit the control arm with an exact GP,
and the treated arm with any scalable surrogate --- a sparse /
inducing-point GP \citep{rasmussen2006gaussian}, or a frequentist
learner with a delta-method variance. \S\ref{sec:experiments}
confirms that this is safe: capping the treated GP at $80$ of
$500$ units leaves coverage at $0.97$, unchanged from the full
fit. The calibrated-uncertainty machinery is essential only where
the data are scarce --- the control arm --- and there it is also
inexpensive.

Two caveats about \GPCATE{}'s calibration are worth stating in
advance of the experiments. First, when $\mu_w$ is smooth and
$d$ small, the empirical-Bayes GP is slightly cautious and
coverage runs a little above nominal ($0.97$--$0.99$ in
\S\ref{sec:experiments}); this is the safe direction. Second,
with few controls in a high-dimensional covariate space,
$s_0(x)$ approaches the prior scale over most of the space,
because the data genuinely under-determines $\mu_0$ there; the
credible interval is then wide. We regard this as a property of the method, not a deficiency:
it is the correct statement that thirty controls in twenty-five
dimensions cannot determine the placebo surface, and
\S\ref{sec:experiments} shows that the alternatives' narrower
intervals in that setting do not cover. \GPCATE{} reports the
difficulty of the inference problem rather than concealing it.

\section{Experiments}
\label{sec:experiments}

\begin{figure}[t]
\centering
\begin{tikzpicture}
\begin{axis}[
  ybar, bar width=11pt, width=0.86\linewidth, height=5.4cm,
  symbolic x coords={linear, non-linear, IHDP},
  xtick=data, ymin=0.33, ymax=1.03,
  ytick={0.4,0.5,0.6,0.7,0.8,0.9,1.0},
  ylabel={coverage (averaged over $\Nzero$)},
  legend style={at={(0.5,-0.22)}, anchor=north, legend columns=3,
                /tikz/every even column/.append style={column sep=6pt}},
  legend cell align=left,
  enlarge x limits=0.32, grid=major, grid style={gray!20},
  font=\small,
]
\addplot[fill=brown!55] coordinates {(linear,0.792) (non-linear,0.614) (IHDP,0.396)};
\addplot[fill=teal!55] coordinates {(linear,0.931) (non-linear,0.822) (IHDP,0.701)};
\addplot[fill=blue!55] coordinates {(linear,0.981) (non-linear,0.945) (IHDP,0.955)};
\draw[dashed,thick] (axis cs:linear,0.95) -- (axis cs:IHDP,0.95);
\legend{Causal Forest, BART, \textsc{GP-CATE}}
\end{axis}
\end{tikzpicture}
\caption{Coverage of nominal-$95\,\%$ intervals, averaged over the
  $\Nzero$ sweep, on the three designs. \GPCATE{} (blue) stays at
  or above the nominal line (dashed); the two standard interval
  estimators --- Causal Forest and BART --- under-cover, most on
  IHDP.}
\label{fig:coverage}
\end{figure}
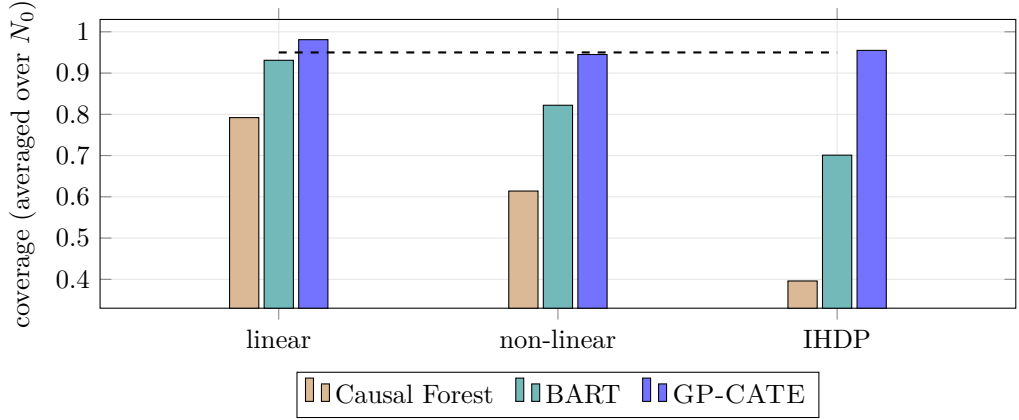

The experiments follow the paper's three claims, in order: the
X-Learner under-covers and the cause is bias
(Table~\ref{tab:bias}); the orthogonal score does not repair it
(Table~\ref{tab:negative}); and \GPCATE{} is calibrated where
the standard interval estimators are not, across three designs
(Table~\ref{tab:coverage}, Figure~\ref{fig:coverage},
Table~\ref{tab:ihdp}).

Coverage requires a known $\tau(x)$, so we evaluate on three
designs. The \emph{linear} design has
$X\sim\mathcal{N}(0,I_4)$,
$W\mid X\sim\mathrm{Bernoulli}(\mathrm{logistic}(0.4X_0))$,
$Y=0.5X_0+0.3X_1+\mathcal{N}(0,0.5^2)+W\tau(X)$ with the linear
$\tau=1+0.3X_0-0.2X_2$. The \emph{non-linear} design has a
two-covariate propensity, an interaction-laden response surface,
and the non-linear $\tau=1+\sin(1.4X_0)-0.3X_2^2$. Finally,
\emph{IHDP} is the semi-synthetic benchmark of
\citet{hill2011bayesian} --- $25$ real covariates and a simulated
response surface with known per-unit $\tau$ --- which we put into
the few-placebo regime by keeping all treated units and
sub-sampling the control arm. The synthetic designs fix
$\None=500$ and sweep $\Nzero$; IHDP sweeps the sub-sampled
$\Nzero$. Full details are in Appendix~\ref{app:repro}.

We compare three families of estimators. The diagnosis and the
negative result concern the standard X-Learner
\citep{kunzel2019metalearners} and the doubly-robust score /
DR-Learner \citep{kennedy2020optimal}; for these we measure the
systematic bias each pseudo-outcome carries. The calibration
comparison is against the two most common CATE estimators that
provide uncertainty intervals: \emph{Causal Forest} --- the
causal-forest DML estimator with honest confidence intervals
\citep{wager2018estimation,athey2016recursive} --- and
\emph{BART}, an S-learner Bayesian additive regression tree whose
CATE posterior is that of $f(x,1)-f(x,0)$
\citep{hill2011bayesian}. For point accuracy we additionally
report the S-, T-, X-, and DR-Learner. No post-hoc calibration is
applied to any method.

With the designs and estimators in place, we now report the
results in the order of the paper's three claims. We turn first
to the diagnosis. Table~\ref{tab:bias} decomposes, over seeds,
the error of a Bayesian regression on the X-Learner
pseudo-outcome --- the
natural way to equip the X-Learner with an interval
(\S\ref{sec:diagnosis}) --- into a systematic and a sampling
component. The systematic bias is a large fraction of the
posterior spread at small $\Nzero$ and decays as the placebo arm
grows, the behaviour (\ref{eq:diag-biased-target}) predicts. A
posterior displaced by $0.55$ standard deviations cannot cover:
the nominal-$95\,\%$ interval covers $0.88$ at $\Nzero{=}30$,
recovering to $0.97$ by $\Nzero{=}500$ as the bias decays.

\begin{table}[h]
\centering\small
\caption{Bias/spread decomposition of a Bayesian regression on the
  X-Learner pseudo-outcome (linear design), with the resulting
  credible-interval coverage. The systematic bias is a large
  fraction of the posterior spread at $\Nzero{=}30$ and the
  interval under-covers.}
\label{tab:bias}
\begin{tabular}{lrrr}
\toprule
 & systematic bias (RMSE) & posterior spread & coverage \\
\midrule
$\Nzero = 30$  & $0.146$ & $0.268$ & $0.88$ \\
$\Nzero = 500$ & $0.028$ & $0.108$ & $0.97$ \\
\bottomrule
\end{tabular}
\end{table}

Next, Table~\ref{tab:negative} measures the negative result of
\S\ref{sec:negative}. The doubly-robust score's systematic bias
is essentially identical to the X-Learner pseudo-outcome's
($0.154$ vs $0.146$ at $\Nzero{=}30$): orthogonalization yields
no improvement here. The bias persists when the propensity is
\emph{known exactly} ($0.174$ at $\Nzero{=}30$) --- the
efficiency/unbiasedness conflict of \S\ref{sec:negative},
not propensity-estimation error --- and recedes only as the
design balances, i.e.\ as overlap is restored. The unweighted
DR-Learner avoids the bias, but its variance is very large:
$\PEHE = 0.63$ at $\Nzero{=}30$, the largest in
Table~\ref{tab:pehe}.

\begin{table}[h]
\centering\small
\caption{Systematic bias of an inverse-variance-weighted
  regression of the doubly-robust score (linear design),
  propensity estimated and known exactly. Neither improves on the
  X-Learner pseudo-outcome's $0.146$ at $\Nzero{=}30$
  (Table~\ref{tab:bias}).}
\label{tab:negative}
\begin{tabular}{lrrr}
\toprule
$\Nzero$ & $30$ & $100$ & $500$ \\
\midrule
DR score, estimated propensity      & $0.154$ & $0.099$ & $0.049$ \\
DR score, propensity known exactly  & $0.174$ & $0.086$ & $0.033$ \\

\bottomrule
\end{tabular}
\end{table}

We now turn to the main calibration result.
Table~\ref{tab:coverage} and Figure~\ref{fig:coverage} compare
\GPCATE{} against the two standard interval estimators. \GPCATE{} is the only estimator we test that is
calibrated throughout: coverage $0.94$--$0.99$ on every design
and sample size. The two most common interval-producing CATE
estimators perform poorly in this regime. \emph{Causal Forest}'s
honest confidence intervals under-cover substantially ---
$0.59$--$0.83$ on the synthetic designs and $0.34$--$0.45$ on
IHDP. \emph{BART} is calibrated on the linear design
($0.91$--$0.94$) but under-covers on the non-linear
($0.79$--$0.86$) and IHDP ($0.59$--$0.77$) designs. \GPCATE{} is
mildly conservative on the smooth linear design
($0.97$--$0.99$) --- the safe direction, as
\S\ref{sec:method} anticipates --- and essentially nominal on the
non-linear and IHDP designs.

\begin{table}[h]
\centering\small
\caption{Coverage of nominal-$95\,\%$ credible intervals, no
  post-hoc calibration. Columns are small/mid/large $\Nzero$ per
  design (linear and non-linear: $30/100/500$; IHDP:
  $30/60/100$). \GPCATE{} is calibrated throughout; both standard
  interval estimators under-cover on the harder designs.}
\label{tab:coverage}
\begin{tabular}{lrrrrrrrrr}
\toprule
& \multicolumn{3}{c}{linear} & \multicolumn{3}{c}{non-linear}
& \multicolumn{3}{c}{IHDP} \\
\cmidrule(lr){2-4}\cmidrule(lr){5-7}\cmidrule(lr){8-10}
method & s & m & l & s & m & l & s & m & l \\
\midrule
Causal Forest \citep{wager2018estimation} & $0.79$ & $0.76$ & $0.83$ & $0.59$ & $0.61$ & $0.63$ & $0.45$ & $0.40$ & $0.34$ \\
BART \citep{hill2011bayesian} & $0.94$ & $0.94$ & $0.91$ & $0.79$ & $0.86$ & $0.82$ & $0.59$ & $0.74$ & $0.77$ \\
\midrule
\textbf{\GPCATE{} (ours)} & $\mathbf{0.97}$ & $\mathbf{0.99}$ & $\mathbf{0.98}$ & $\mathbf{0.94}$ & $\mathbf{0.94}$ & $\mathbf{0.95}$ & $\mathbf{0.96}$ & $\mathbf{0.96}$ & $\mathbf{0.95}$ \\
\bottomrule
\end{tabular}
\end{table}

Among the three designs, IHDP warrants the closest examination:
with $25$ real covariates and few controls, it is the most
demanding test.
Table~\ref{tab:ihdp} pairs coverage with interval width.
\GPCATE{} is the only calibrated estimator ($0.95$--$0.96$), and
its intervals are wide ($\approx 12$--$18$). This width is the
accurate report: with thirty controls in twenty-five dimensions
the placebo surface is near-undetermined, and the GP posterior
reflects this. It is \emph{not} a vague prior reasserting itself
--- the estimator is performing valid inference, not abdicating.
The control-arm GP posterior variance contracts to $0.55$ of its
(proper, fitted) prior at $\Nzero{=}30$ and to $0.16$ at
$\Nzero{=}100$; on the lower-dimensional linear and non-linear
designs the contraction reaches $0.05$--$0.07$
(Appendix~\ref{app:repro}). The GP extracts useful information
from the few controls, but in twenty-five dimensions cannot
extract enough to make the interval narrow. Causal Forest and
BART report intervals a third to a quarter as wide and
under-cover substantially --- Causal Forest's $3$--$4$-wide
intervals cover only $0.34$--$0.45$. Their intervals are narrow
precisely because they are miscalibrated. A practitioner is
better served by \GPCATE{}'s wide, calibrated interval, which
indicates that more control data or a lower-dimensional analysis
is required, than by a narrow interval that miscovers a third of
the time.

\begin{table}[h]
\centering\small
\caption{IHDP: coverage and mean interval width, $\Nzero$
  sub-sampled. \GPCATE{} covers; both standard interval
  estimators under-cover, at a third to a quarter of the width.}
\label{tab:ihdp}
\begin{tabular}{lrrr rrr}
\toprule
& \multicolumn{3}{c}{coverage (target $0.95$)} & \multicolumn{3}{c}{mean width} \\
\cmidrule(lr){2-4}\cmidrule(lr){5-7}
$\Nzero$ & $30$ & $60$ & $100$ & $30$ & $60$ & $100$ \\
\midrule
Causal Forest             & $0.45$ & $0.40$ & $0.34$ & $4.0$ & $3.5$ & $3.0$ \\
BART                      & $0.59$ & $0.74$ & $0.77$ & $3.8$ & $3.8$ & $3.7$ \\
\midrule
\textbf{\GPCATE{} (ours)} & $\mathbf{0.96}$ & $\mathbf{0.96}$ & $\mathbf{0.95}$ & $15.5$ & $17.9$ & $12.2$ \\
\bottomrule
\end{tabular}
\end{table}

Beyond coverage and width, we also assess point accuracy:
calibration is not achieved at the expense of estimation error.
Table~\ref{tab:pehe} reports $\PEHE$ on the linear design: \GPCATE{} is the most
accurate estimator at $\Nzero \ge 100$ and competitive at
$\Nzero{=}30$, where BART edges it. On the non-linear design
\GPCATE{}'s $\PEHE$ is $0.51/0.33/0.25$ across the sweep, on a par
with BART ($0.51/0.38/0.28$) and ahead of Causal Forest
($0.65/0.49/0.39$). On IHDP, BART is in fact the most accurate
\emph{point} estimator ($\PEHE \approx 2.4$--$3.7$ versus
\GPCATE{}'s $3.5$--$4.5$) --- but its intervals under-cover
(Table~\ref{tab:ihdp}). In summary, \GPCATE{} is competitive on
accuracy across all designs and is the only method that is also
calibrated; where a standard estimator is more accurate, it is
so while reporting intervals that do not cover.

\begin{table}[h]
\centering\small
\caption{$\PEHE$ on the linear design, $\None=500$ (lower is
  better). \GPCATE{} is the most accurate at $\Nzero\ge100$.}
\label{tab:pehe}
\begin{tabular}{lrrr}
\toprule
$\Nzero$ & $30$ & $100$ & $500$ \\
\midrule
S-Learner                       & $0.36$ & $0.24$ & $0.18$ \\
T-Learner                       & $0.46$ & $0.37$ & $0.27$ \\
X-Learner                       & $0.58$ & $0.23$ & $0.11$ \\
DR-Learner (unweighted)         & $0.63$ & $0.49$ & $0.18$ \\
Causal Forest                   & $0.31$ & $0.24$ & $0.16$ \\
BART                            & $\mathbf{0.23}$ & $0.20$ & $0.15$ \\
\textbf{\GPCATE{} (ours)}       & $0.29$ & $\mathbf{0.13}$ & $\mathbf{0.07}$ \\
\bottomrule
\end{tabular}
\end{table}

Taken together, these experiments establish the paper's three
claims. The X-Learner under-covers in the few-placebo regime
from first-order nuisance bias (Table~\ref{tab:bias}); the
orthogonal-score fix does not remove that bias, even with the
propensity known (Table~\ref{tab:negative}). \GPCATE{}, which
assigns the hard-to-learn nuisance a full Bayesian posterior
rather than a point estimate, is calibrated across the linear,
non-linear, and
IHDP designs (Table~\ref{tab:coverage}) and accurate
(Table~\ref{tab:pehe}) --- the only calibrated method we test,
the standard Causal Forest and BART included. Its cost is
interval width on high-dimensional covariates
(Table~\ref{tab:ihdp}), reflecting the true difficulty of the
problem rather than concealing it.

\section{Discussion}
\label{sec:discussion}

In the few-placebo regime the practical choice of estimator is
governed by calibration, and \GPCATE{} is the only estimator we
tested that is calibrated across all three designs
(\S\ref{sec:experiments}). The standard interval-producing
estimators each fail somewhere: Causal Forest under-covers
severely, down to $0.34$ on IHDP --- a misplaced interval, the
dangerous direction --- and BART, calibrated on the smooth linear
design, under-covers once the response surface is non-linear or
high-dimensional. We therefore recommend \GPCATE{} as the default
when one arm is genuinely small; once the arms approach balance
the few-placebo problem dissolves and a plain X-Learner is
adequate and cheaper.

However, \GPCATE{}'s intervals on the $25$-dimensional IHDP
covariates are wide --- several times the spread of $\tau$
itself. We have not treated this as a defect to be engineered
away. Thirty control units in twenty-five dimensions do not
determine the placebo surface, and a calibrated interval must
reflect this; the alternatives' narrower IHDP intervals are
narrower precisely because they under-cover
(\S\ref{sec:experiments}). A practitioner is better served by a
wide, calibrated interval than by a narrow one that miscovers:
the wide interval correctly indicates that more control data, or
a lower-dimensional analysis, are required before a confident
claim. The width is, however, a real limitation on what can be
concluded from such data, and reducing it without sacrificing
coverage --- by borrowing strength across arms with a coupled GP
\citep{alaa2017bayesian}, or by a data-driven dimension reduction
--- is the most useful direction for future work.

More generally, the diagnosis and the negative result share a
principle that extends beyond the few-placebo regime: a variance
attached to a point estimate of a hard-to-learn nuisance cannot
be expected to yield calibration. A point estimate's regularization bias is invisible
to any variance computed around it, whether that variance comes
from a homoskedastic likelihood, an errors-in-variables
weighting, an orthogonal score, or an explicit bias model. The
reliable route to a calibrated downstream interval is a nuisance
that carries its own posterior. \GPCATE{} is one instantiation;
the principle is broader.

Several strands of related work bear on this picture. The
X-Learner is \citet{kunzel2019metalearners}; the doubly-robust
score and DR-Learner are \citet{robins1995semiparametric} and
\citet{kennedy2020optimal}; cross-fitting and the orthogonality
framework are \citet{chernozhukov2018double}. That orthogonal
estimators degrade under poor overlap is known for the average
effect; our contribution is to show the few-placebo regime
\emph{is} a poor-overlap regime by construction, and to add the
propensity-independent failure of \S\ref{sec:negative}.
Gaussian-process and Bayesian-nonparametric models for treatment
effects are established --- the multi-task GP of
\citet{alaa2017bayesian}, and BART-based BCF
\citep{hahn2020bayesian}; \GPCATE{} is a deliberately plain
instance, and our point is diagnostic rather than architectural:
a fully-Bayesian nuisance is what the few-placebo regime
requires. We benchmark directly against the two most common
interval-producing CATE estimators --- Causal Forest
\citep{wager2018estimation,athey2016recursive} and BART
\citep{hill2011bayesian} --- and find both under-cover in this
regime (\S\ref{sec:experiments}): Causal Forest's honest
asymptotic intervals and BART's posterior are calibrated under
good overlap but not when one arm is small. Generalized-Bayes
and bootstrap calibrations
\citep{lyddon2019general,linhan2026bootstrap,javurek2026generalized}
and semiparametric Bernstein--von Mises results
\citep{rayvandervaart2020semiparametric,breunig2025double} target
settings with good overlap; conformal meta-learners
\citep{alaa2023conformal,leicandes2021conformal} give
finite-sample intervals for the individual effect rather than a
posterior for the CATE function.

Four limitations of \GPCATE{} merit explicit mention. First, an
exact GP costs $\mathcal{O}(N_w^3)$ per arm. The control arm,
small by definition, is inexpensive; for a large treated arm
\S\ref{sec:method} provides a concrete prescription --- a sparse
surrogate, justified because the treated arm carries only
$6$--$25\,\%$ of the CATE variance --- so this is a resolved
engineering choice, not an obstacle to practical use. Second,
\GPCATE{} is mildly conservative on smooth, low-dimensional
designs (\S\ref{sec:experiments}); the safe direction, but a
mild efficiency loss. Third, in high dimension with very few
controls the calibrated interval can be wide enough to limit
what the data support --- a property of the problem, made
explicit rather than concealed. Fourth, the kernel family is a
modelling choice; we use an isotropic squared-exponential, since
per-dimension length scales over-fit from a small control arm
(\S\ref{sec:experiments}).

\section{Conclusion}
\label{sec:conclusion}

In the few-placebo regime, the X-Learner with a Bayesian second
stage produces credible intervals that under-cover. The cause is
structural: the X-Learner pseudo-outcome inherits the bias of
the nuisance model fitted to the small control arm, and a
posterior centered on a biased target is displaced rather than
merely widened. The standard correction --- an orthogonal
doubly-robust score --- does not resolve this, because the
few-placebo regime is itself a poor-overlap regime: the score's
nominally second-order remainder is amplified there, and the
efficient inverse-variance-weighted variant trades that
amplification back for the bias. Bias and variance cannot be
controlled jointly through orthogonalization alone.

Both failures share a single cause. Each attaches a variance to a
\emph{point estimate} of the placebo nuisance, and a point
estimate's regularization bias is invisible to any variance
computed around it. \GPCATE{} replaces the point estimate with a
per-arm Gaussian process, so the small arm's uncertainty enters
the CATE posterior directly: the posterior mean and the posterior
variance are two aspects of a single coherent object, rather
than a point estimate with a separately attached error bar. The
CATE posterior is the difference of the two arm posteriors, and
the construction requires no cross-fitting, no orthogonal score,
and no auxiliary bias model.

Across a linear design, a non-linear design, and the IHDP
semi-synthetic benchmark, \GPCATE{} is the only estimator we
evaluate that attains calibrated coverage throughout; Causal
Forest, BART, and the standard meta-learners do not. The cost is
interval width: when few controls do not determine the placebo
surface in high dimension, \GPCATE{}'s intervals widen
accordingly --- the accurate answer that the data support, in
contrast to the narrower but miscalibrated intervals of the
alternatives.

\bibliographystyle{abbrvnat}
\bibliography{refs}

\appendix
\section{Derivation of the Doubly-Robust Remainder}
\label{app:remainder}

We derive equation~(\ref{eq:neg-remainder}). Fix the cross-fitted
nuisances $\hat\mu_0,\hat\mu_1,\hat\pi$ and take expectations over
$(W,Y)$ given $X$. With $\E[W\mid X]=\pi$ and
$\E[Y\mid X,W{=}w]=\mu_w$,
\begin{align*}
  \E\!\left[\tfrac{W(Y-\hat\mu_1)}{\hat\pi}\,\Big|\,X\right]
  &= \tfrac{\pi}{\hat\pi}(\mu_1-\hat\mu_1),
  &
  \E\!\left[\tfrac{(1-W)(Y-\hat\mu_0)}{1-\hat\pi}\,\Big|\,X\right]
  &= \tfrac{1-\pi}{1-\hat\pi}(\mu_0-\hat\mu_0).
\end{align*}
Substituting into (\ref{eq:bg-dr}) and using
$\hat\mu_w = \mu_w - (\mu_w-\hat\mu_w)$,
\begin{align*}
  \E[D^{\mathrm{DR}}\mid X]
  &= (\mu_1-\mu_0)
   + \Bigl(\tfrac{\pi}{\hat\pi}-1\Bigr)(\mu_1-\hat\mu_1)
   - \Bigl(\tfrac{1-\pi}{1-\hat\pi}-1\Bigr)(\mu_0-\hat\mu_0).
\end{align*}
The first term is $\tau(X)$; the rest is $R(X)$ of
(\ref{eq:neg-remainder}), each summand a product of a propensity
error and an outcome-nuisance error --- the second-order
property. The arm-conditional means (\ref{eq:neg-conditional})
follow by not averaging over $W$. The first-order expansion
$\tfrac{1-\pi}{1-\hat\pi}-1 = \tfrac{\hat\pi-\pi}{1-\hat\pi}
\approx \tfrac{\hat\pi-\pi}{1-\pi}$ holds for
$|\hat\pi-\pi|\ll 1-\pi$ and gives (\ref{eq:neg-amplified}); the
$1/(1-\pi)$ factor is the overlap amplification.

\section{Implementation and Reproducibility}
\label{app:repro}

The three designs are specified as follows. The \emph{linear}
design uses $X\sim\mathcal{N}(0,I_4)$,
$W\mid X\sim\mathrm{Bernoulli}(\mathrm{logistic}(0.4X_0))$,
$Y=0.5X_0+0.3X_1+0.5\xi+W\tau(X)$ with $\tau=1+0.3X_0-0.2X_2$.
The \emph{non-linear} design has a two-covariate propensity, an
interaction-laden response surface, and a non-linear
$\tau=1+\sin(1.4X_0)-0.3X_2^2$. \emph{IHDP} is the semi-synthetic
benchmark of \citet{hill2011bayesian} --- $25$ real covariates,
$747$ units, and a simulated response surface with known per-unit
$\tau$; we keep all treated units and sub-sample the control arm
to $\Nzero$ to mirror $\Nzero\ll\None$. The few-placebo regime
in the synthetic designs is produced by stratified sampling to
$\Nzero$ controls and $\None{=}500$ treated; the known-propensity
check of \S\ref{sec:experiments} draws
$W\sim\mathrm{Bernoulli}(p)$ with $p$ fixed, so $\pi\equiv p$.

For the estimators themselves, \GPCATE{} fits, per arm, a
Gaussian process with a constant-times-RBF-plus-white-noise
kernel, hyperparameters by marginal likelihood with two restarts;
the RBF length scale is isotropic (per-dimension ARD length
scales over-fit a small control arm and \emph{worsen} coverage
--- on IHDP, to $0.78$). The bias measurements of
\S\ref{sec:experiments} regress the
X-Learner pseudo-outcome and the doubly-robust score, with
gradient-boosted nuisances (depth $3$, $200$ estimators) and
$K{=}5$ cross-fitting. The interval baselines are the
\texttt{econml} causal-forest DML estimator ($500$ trees,
gradient-boosted nuisance models, honest confidence intervals via
\texttt{effect\_interval}) and an S-learner BART on $[X,W]$ ($60$
trees, $200$ posterior draws after $100$ burn-in), its CATE
interval taken from the posterior of $f(x,1)-f(x,0)$; the S-, T-,
X-, and DR-Learner use the same gradient-boosted nuisances.

Synthetic-design numbers average $20$ seeds: coverage, $\PEHE$,
and width on $1{,}500$ fresh test points, and systematic bias as
the RMSE of the seed-averaged prediction against $\tau$ on a
fixed $3{,}000$-point set. IHDP averages the $10$ response-surface
replications, each with $4$ control sub-samples, with coverage
evaluated against the known per-unit $\tau$ at all $747$
covariate points.

The posterior-contraction ratios reported in
\S\ref{sec:experiments} and the treated-arm variance share
discussed in \S\ref{sec:method} compare,
per arm, the fitted prior function variance --- the
\texttt{ConstantKernel} amplitude $\sigma_w^2$ --- with the mean
posterior function variance $s_w(x)^2$ over the test set; the
contraction ratio is their quotient. The contraction is
$0.05$--$0.07$ on the linear and non-linear designs and $0.55$
($\Nzero{=}30$) to $0.16$ ($\Nzero{=}100$) on IHDP; the treated
arm's share $s_1^2/(s_0^2+s_1^2)$ is $0.06$--$0.25$ in the
few-placebo regime. As a sanity check, we also fit the treated GP
on a random subset of the treated arm; coverage on the linear
design at $\Nzero{=}30$ is $0.97$ for caps of $500$, $200$, and
$80$.

\end{document}